\ifdefined\pdfoutput
  \pdfoutput=1
\fi
\documentclass[10pt,twocolumn]{article}

\usepackage[letterpaper,margin=0.75in]{geometry}
\usepackage{mathtools,amssymb}
\usepackage[hidelinks]{hyperref}
\usepackage{graphicx}
\usepackage{booktabs}
\usepackage{flushend}
\usepackage[numbers,sort&compress]{natbib}
\usepackage{microtype}
\usepackage{abstract}
\usepackage[hidelinks]{hyperref}

\hypersetup{
  pdftitle={CVE-SAI: Counterfactual Visual Evidence-Guided Selective Attribute Indexing for Risk-Controlled E-commerce Search},
  pdfauthor={Author Name(s)},
  pdfsubject={Risk-controlled selective attribute indexing},
  pdfkeywords={Product Attribute Indexing, Multimodal Retrieval, Risk Control, E-commerce Search}
}

\AtBeginDocument{%
  }

\title{CVE-SAI: Counterfactual Visual Evidence-Guided Selective Attribute Indexing for Risk-Controlled E-commerce Search}
\author{%
Xiaolong Sun\(^{1}\)\quad
Qichao Wang\(^{2}\)\quad
Hangyu Li\(^{3}\)\quad
Liang Chen\(^{1,*}\)\\[3pt]
\(^{1}\)Sun Yat-Sen University, Guangzhou, China\\
\(^{2}\)Nanyang Technological University, Singapore\\
\(^{3}\)Tencent, Shenzhen, China\\[3pt]
\texttt{sunxlong@mail2.sysu.edu.cn}\quad
\texttt{qichao001@e.ntu.edu.sg}\\
\texttt{masonhyli@tencent.com}\quad
\texttt{chenliang6@mail.sysu.edu.cn}\\
\(^{*}\)Corresponding author
}
\date{}

\begin{document}

\twocolumn[
\maketitle
\begin{onecolabstract}

Multimodal product models can complete missing e-commerce attributes, yet
current methods still optimize attribute-answer accuracy without verifying
visual support, conflate transient prediction with persistent index admission,
and lack explicit risk control over factually incorrect or visually
unsupported values. We address these gaps with Counterfactual Visual
Evidence-Guided Selective Attribute Indexing (CVE-SAI), which first infers and
freezes an ontology-constrained candidate from the primary image and attribute
question without catalog text, and then decides whether that candidate should
enter the index. Focus-Zone Distortion (FZD) constructs an attribute-specific
visual-dependence proxy through a controlled counterfactual intervention, and
Evidence-Guided Attention Redistribution (EGAR) uses the proxy to refine
ontology-constrained scoring. The canonical candidate is frozen before
evidence necessity, evidence retention, nuisance-transformation stability, and
candidate-specific catalog-text conflict audits; catalog text can only tighten
admission and cannot revise the candidate. Independent family-level
calibration selects one policy with a simultaneous one-sided finite-sample
bound under a 5\% unsafe-admission budget. Experiments on five visual
attributes derived from Amazon Berkeley Objects show that CVE-SAI improves
attribute inference and evidence localization, achieves the highest certified
admission coverage under the shared risk protocol, and yields the strongest
controlled retrieval performance with the lowest unsafe auto-induced exposure
among automatic-admission systems. Separating inference from admission
therefore enables visually supported attribute completion to improve retrieval
while limiting persistent index contamination.
\end{onecolabstract}

\vspace{0.75em}

]

\section{Introduction}
\label{sec:introduction}

Product attributes support product retrieval, ranking, recommendation, and catalog understanding, yet large catalogs often leave attribute values missing or incomplete. MAVE~\cite{Yang2022MAVE} documents this problem across diverse product categories, and MOON~\cite{Zhang2026MOON} shows that visual and textual product content can support attribute prediction and cross-modal retrieval. Product imagery therefore offers a practical source for completing catalog records. The same prediction becomes consequential when it is stored and repeatedly reused by search infrastructure rather than consumed once.

Multimodal large language models can infer product attributes from images and jointly represent heterogeneous product content. Their fluency does not ensure that every output is visually supported. Visual Contrastive Decoding (VCD)~\cite{Leng2024VisualContrastiveDecoding} mitigates unsupported generation by contrasting outputs from clean and distorted images. For product attributes, language-model confidence, ontology validity, and support from the primary image are distinct properties. A candidate may satisfy the first two and still be unsuitable for a persistent catalog update.

Existing work addresses either product attribute inference and representation learning or visually unsupported multimodal generation. Neither direction jointly verifies support from the current product image and controls whether a prediction enters a persistent index. A visually plausible answer may therefore remain unsuitable as a durable search signal.

A one-time prediction error becomes persistent retrieval contamination once the inferred attribute is indexed and reused for matching and ranking. A suitable admission rule should respond to changes in attribute-relevant evidence while remaining stable under semantics-preserving nuisance variation, mirroring the counterfactual behavior studied in robust retrieval~\cite{Chen2026CounterfactualRetrieval}. This distinction calls for an index-admission decision that is separate from attribute inference.

We formulate risk-controlled selective product attribute indexing as a decision separate from attribute inference. The model may answer or abstain, but only a frozen canonical value can be considered for admission. Image truth records what the primary image supports, whereas item truth records the trusted product value. SelectiveNet~\cite{Geifman2019SelectiveNet} motivates the reject option and risk--coverage trade-off, while Learn then Test~\cite{Angelopoulos2025LearnThenTest} supports policy selection under a prespecified risk budget. We set the unsafe-admission budget to 5\% on an independent family-level calibration population.

As illustrated in Figure~\ref{fig:framework}, we propose Counterfactual Visual Evidence-Guided Selective Attribute Indexing (CVE-SAI), which separates product-text-free attribute inference from risk-controlled index admission. Focus-Zone Distortion (FZD) constructs an attribute-specific visual-dependence proxy, and Evidence-Guided Attention Redistribution (EGAR) uses it to refine scores over the ontology. CVE-SAI freezes the resulting canonical candidate before necessity, retention, nuisance-transformation stability, and candidate-specific catalog-text conflict audits. Independent calibration then determines whether the candidate enters the isolated auto-attribute field.

We construct a five-attribute benchmark and controlled retrieval evaluation from Amazon Berkeley Objects (ABO)~\cite{Collins2022ABO}. Across the five visually auditable attributes, CVE-SAI achieves the highest Macro Visual Answer Accuracy (Macro VAA) and Answerability Macro-F1 (Ans.-F1) among the evaluated visual-answering systems, while FZD obtains the highest Macro Patch AUPRC among evidence-localization methods. Under the prespecified 5\% unsafe-admission budget, CVE-SAI attains the highest Certified Write Coverage (CWC@5\%) among the evaluated risk-controlled systems. Its admitted attributes also yield the highest NDCG@10 and the lowest Unsafe Auto-Induced Exposure@10 (UAIE@10) among automatic-admission systems in controlled Lucene retrieval. Our contributions are as follows:
\begin{itemize}
    \item We formulate risk-controlled selective product attribute indexing, which separates attribute inference from persistent index admission. The formulation distinguishes image truth from item truth, supports explicit abstention, and defines unsafe-admission risk for admitted canonical values.

    \item We introduce the visual front end of CVE-SAI. FZD derives an attribute-specific visual-dependence proxy from a controlled counterfactual edit, and EGAR uses this proxy to improve ontology-constrained candidate scoring. The frozen candidate is then examined through evidence necessity, evidence retention, nuisance-transformation stability, and candidate-specific catalog-text conflict audits.

    \item We develop a risk-controlled admission strategy that uses independent finite-sample calibration to select the feasible policy with the highest coverage under a prespecified budget for unsafe admission. The selected policy is frozen before test, and only admitted canonical values enter the isolated auto-attribute field.

    \item We construct a five-attribute benchmark from ABO and evaluate CVE-SAI across attribute inference, evidence localization, index admission, and retrieval with a fixed Lucene pipeline. CVE-SAI achieves the highest certified admission coverage under the 5\% risk budget, improves retrieval effectiveness, and reduces unsafe auto-induced exposure among the evaluated automatic-admission systems.
\end{itemize}

\section{Related Work}
\label{sec:related_work}

\subsection{Product Attribute Extraction}

Product attributes support product ranking, retrieval, and recommendation, motivating benchmarks such as MAVE, which combines titles, descriptions, features, and specifications across diverse categories and exposes incomplete values and a challenging zero-shot split. MBSD~\cite{Liu2023MBSD} transfers knowledge across product modalities through self-distillation, whereas MOON learns generative multimodal representations for attribute prediction and cross-modal retrieval. These representation-learning approaches predict or encode product content but do not decide whether inferred attributes should enter a persistent index. MXT~\cite{Khandelwal2023MXT} formulates large-scale multimodal extraction as question answering over images and text, while DEFLATE~\cite{Zhang2023DEFLATE} generates explicit and implicit values before assessing candidate credibility with a discriminator. EIVEN~\cite{Zou2024EIVEN} efficiently adapts multimodal language models for implicit values, and ImplicitAVE~\cite{Zou2024ImplicitAVE} provides a curated benchmark for this setting. Under image-only inference, ViOC-AG~\cite{Gong2025ViOCAG} transfers textual attribute knowledge to a visual generator and corrects out-of-domain values, whereas MICE~\cite{Gong2025MICE} combines specialized captioning experts to provide complementary visual descriptions. HyperPAVE~\cite{Gong2024MultiLabelPAVE} models higher-order relations induced by user behavior and product inventory in a heterogeneous hypergraph and uses inductive link prediction for unseen values, while Hypergraph PAVE~\cite{Hu2025HypergraphPAVE} integrates visual and textual product information into multimodal hypergraphs for zero-shot attribute prediction. MSIT~\cite{Li2025MSIT} extends attribute mining to the open world through multimodal self-correction, and TACLR~\cite{Su2025TACLR} uses taxonomy-aware retrieval to handle implicit and out-of-distribution values while producing normalized outputs. Beyond product attributes, HADSF~\cite{Nie2026HADSF} uses structured aspect--opinion signals to improve downstream recommendation, illustrating the value of evaluating extracted information through its system-level effect. These methods improve attribute generation, extraction, correction, or normalization, but generally treat prediction as the endpoint. CVE-SAI instead separates product-text-free candidate generation from risk-controlled persistent index admission.

\subsection{Reliable Multimodal Prediction}

POPE~\cite{Li2023POPE} evaluates object hallucination through binary questions about whether mentioned objects are present in an image rather than scoring unconstrained free-form descriptions. Visual Contrastive Decoding contrasts outputs from clean and distorted images, while M3ID~\cite{Favero2024M3ID} strengthens dependence on the visual prompt through mutual-information decoding. These decoding methods reduce unsupported content without retraining the base model. Their interventions affect current responses rather than persistent catalog state. AGLA~\cite{An2025AGLA} assembles global and local attention, whereas CMAC~\cite{Li2026CMAC} calibrates cross-modal attention to balance visual and textual influence during training-free decoding. Same Attention, Different Truths~\cite{Wang2026SameAttention} shows that similar visual-attention magnitudes can correspond to different object-hallucination outcomes, motivating logit-aware diagnosis. VES-RFT~\cite{Hou2026VESRFT} rewards visual-evidence sensitivity during reinforcement fine-tuning, whereas CausalLens~\cite{Ji2026CausalLens} intervenes on sensitivity-selected attention heads. EnAR~\cite{Liang2026EnAR} guides generation with counterfactual visual impressions. Together, these methods motivate measuring visual dependence separately for each attribute under controlled interventions. Along a textual evidence channel, Ext2Gen~\cite{Song2026Ext2Gen} extracts evidence before generation. KnowFC~\cite{Zhang2026KnowFC} studies conflicts between external evidence and parametric knowledge. These methods use retrieved text for generation or fact verification, whereas CVE-SAI reads catalog text only after freezing the visual candidate and uses it solely to tighten admission. SelectiveNet integrates a reject option and jointly optimizes prediction and selection. Learn then Test selects predictive procedures through finite-sample tests, while Conformal Risk Control~\cite{Angelopoulos2024ConformalRiskControl} controls monotone losses through calibration. Two-stage Risk Control~\cite{Xu2025TwoStageRisk} separately controls candidate retrieval and ranking quality at query time. These risk-control methods operate on query-time predictions or rankings rather than persistent catalog updates. Counterfactual dense retrieval promotes sensitivity to relevance-bearing changes and stability under irrelevant variation through unsupervised contrastive learning. CVE-SAI introduces FZD as an attribute-specific visual-dependence proxy and EGAR to use that proxy in ontology-constrained scoring. The visual candidate is frozen before necessity, retention, nuisance-transformation stability, and catalog-text conflict audits; catalog text can tighten admission but cannot revise the candidate. Independent finite-sample calibration then selects a policy under the prespecified unsafe-admission budget, and only admitted canonical values enter the isolated auto-attribute field.

\section{Problem Formulation}
\label{sec:problem}

\noindent\textbf{Task and Decision.}
For product--attribute pair $i$, let $I_i$ be the primary product image,
$X_i$ the catalog text, and $a_i$ the target attribute. The frozen ontology is
$\mathcal V_{a_i}=\{v_1,\ldots,v_{m_{a_i}}\}$. The visual-answer space is
$\mathcal C_{a_i}=\mathcal V_{a_i}\cup\{o,\bot\}$, where $o$ denotes an
image-supported value outside the ontology and $\bot$ denotes that the primary
image is insufficient to answer. A pre-split applicability table fixes which
product--attribute pairs enter the visual-answer population; once included,
abstention, an out-of-ontology response, model fallback, technical failure, and
withholding remain in the evaluation denominator.

The visual stage predicts $\hat y_i\in\mathcal C_{a_i}$ using only $I_i$,
$a_i$, and the frozen ontology. An admission policy returns
$d_{\theta,i}\in\{0,1\}$, where $1$ denotes admission and $0$ denotes
withholding. Catalog text is unavailable until the visual candidate is frozen
and may affect only $d_{\theta,i}$. Consequently, only a canonical
$\hat y_i\in\mathcal V_{a_i}$ can enter the auto-attribute field.

\noindent\textbf{Dual Truths and Visual Support.}
Image truth and item truth answer different questions. When image-truth
adjudication is complete, $y_i^{\mathrm{img}}\in\mathcal C_{a_i}$ records what
the primary image supports. When item-truth adjudication is complete,
$y_i^{\mathrm{item}}$ records the product fact using trusted structured fields
and independent auxiliary views; the title, description, and primary image are
excluded from this adjudication. Item truth is canonical, out of ontology, or
indeterminate (the internal \texttt{ITEM\_UNKNOWN} state). For canonical value
$v\in\mathcal V_{a_i}$ with adjudicated evidence mask
$\mathbf G_i(a_i,v)$, visual support is
\begin{equation}
 g_i(a_i,v)=
 \mathbf{1}[y_i^{\mathrm{img}}=v]\,
 \mathbf{1}[\|\mathbf G_i(a_i,v)\|_1>0].
 \label{eq:visual-support}
\end{equation}
A canonical image truth requires a complete nonempty mask for its matching
value. An out-of-ontology or visually unanswerable image truth gives zero
support to every canonical value.

\noindent\textbf{Observable Certification Population.}
Risk certification starts from a prespecified, label-blind family sample.
Before any truth annotation or model output is available, one applicable pair
is fixed for each product family in subset $s$, forming
$\mathcal D_{\mathrm{pre}}^{(s)}$. This selected-pair list remains fixed after
annotation attrition and defines the product family as the certification unit.
Let $L_i=1$ when image truth, item truth, and the required evidence annotation
are complete and item truth is canonical or out of ontology. The
risk-observable population is
\begin{equation}
 \mathcal D_{\mathrm{cert}}^{(s)}=
 \{i\in\mathcal D_{\mathrm{pre}}^{(s)}:L_i=1\}.
 \label{eq:observable-certification-population}
\end{equation}
Incomplete truth, indeterminate item truth, or incomplete evidence annotation
causes unreplaced attrition. Membership in
$\mathcal D_{\mathrm{cert}}^{(s)}$ depends only on label observability;
model fallback, refusal, failure, and withholding remain in its denominator.
The certification analysis assumes independently and identically sampled
family-level units.

\noindent\textbf{Unsafe Admission and Objective.}
For an admitted canonical candidate, factual risk captures disagreement with
the product fact and grounding risk captures the absence of visual support:
\begin{equation}
\begin{aligned}
R_i^{\mathrm{fact}}&=\mathbf{1}[\hat y_i\neq y_i^{\mathrm{item}}],\\
R_i^{\mathrm{ground}}&=\mathbf{1}[g_i(a_i,\hat y_i)=0],\\
R_i^{\mathrm{unsafe}}&=R_i^{\mathrm{fact}}\lor R_i^{\mathrm{ground}}.
\end{aligned}
\label{eq:unsafe-admission}
\end{equation}
The union counts an admission once even when both risks occur. Let
$\mathcal A_{\theta}^{(s)}=\{i\in\mathcal D_{\mathrm{cert}}^{(s)}:
 d_{\theta,i}=1\}$, $N_s=|\mathcal D_{\mathrm{cert}}^{(s)}|$,
$n_{\theta}^{(s)}=|\mathcal A_{\theta}^{(s)}|$, and
$k_{\theta}^{(s)}=\sum_{i\in\mathcal A_{\theta}^{(s)}}R_i^{\mathrm{unsafe}}$.
Coverage and the unsafe fraction among admitted values are
\begin{equation}
\begin{aligned}
\operatorname{Coverage}^{(s)}(\theta)&=\frac{n_{\theta}^{(s)}}{N_s},\\
\operatorname{UnsafeWWR}^{(s)}(\theta)&=
\frac{k_{\theta}^{(s)}}{n_{\theta}^{(s)}}.
\end{aligned}
\label{eq:risk-coverage}
\end{equation}
When no value is admitted, coverage is zero and UnsafeWWR is undefined. Let
$U_{\mathrm{CP}}(\theta)$ be the simultaneous one-sided upper confidence bound
defined in Section~\ref{sec:risk-admission}. CVE-SAI maximizes calibration
coverage under the risk and minimum-admission constraints
\begin{equation}
\begin{aligned}
\max_{\theta\in\Theta}\quad
&\operatorname{Coverage}^{(\mathrm{cal\text{-}risk})}(\theta)\\
\text{s.t.}\quad
&U_{\mathrm{CP}}(\theta)\leq\alpha,\qquad n_\theta\geq n_{\min}.
\end{aligned}
\label{eq:risk-constrained-objective}
\end{equation}
The main protocol fixes $\alpha=0.05$ and $n_{\min}=172$; the policy family
and unique selection rule are specified below.

\begin{figure*}[t]
    \centering
    \includegraphics[width=\linewidth]{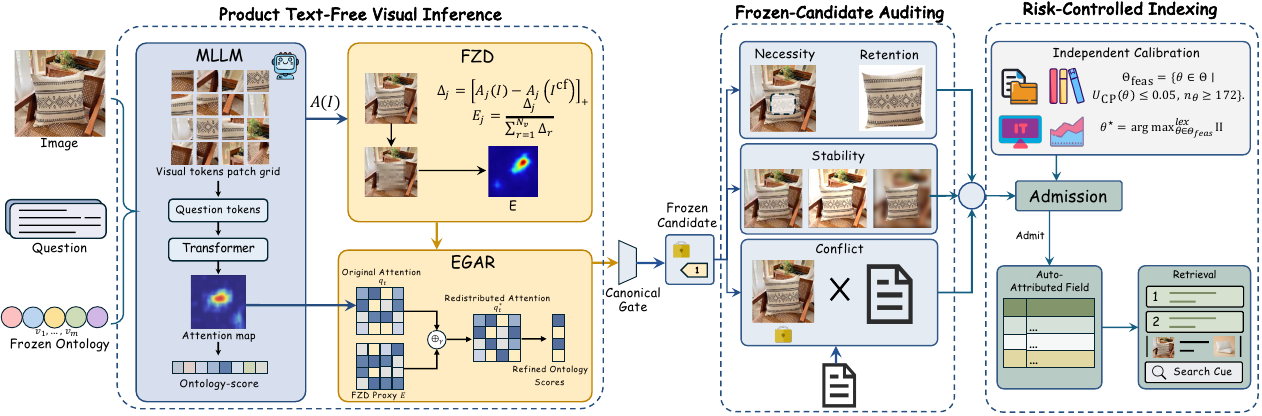}
    \caption{Overview of CVE-SAI. Given a primary image, an attribute question, and a frozen ontology, FZD constructs the attribute-specific visual-dependence proxy $E$, and EGAR redistributes visual attention to refine ontology-constrained scores. The resulting canonical candidate is frozen before evidence necessity, evidence retention, nuisance-transformation stability, and candidate-specific catalog-text conflict audits. Independent calibration selects the admission policy under the 5\% risk budget and the minimum-admission requirement; only admitted values enter the isolated auto-attribute field for retrieval.}
    \label{fig:framework}
\end{figure*}

\section{Methodology}
\label{sec:method}

\subsection{Framework Overview}
\label{sec:method-overview}

Figure~\ref{fig:framework} summarizes three stages: product-text-free visual
inference, frozen-candidate auditing, and independently calibrated index
admission. Given a primary image, an attribute question, and a frozen ontology,
FZD constructs an attribute-specific visual-dependence proxy from a
counterfactually weakened image. EGAR uses that proxy to refine
ontology-constrained scores, after which the canonical gate immediately freezes
the candidate. The auditing block then measures evidence necessity, evidence
retention, nuisance-transformation stability, and candidate-specific
catalog-text conflict. The final block applies the independently selected policy
and writes only admitted values to the isolated auto-attribute field.

Candidate generation and visual auditing use the primary image, attribute
question, ontology, and frozen verbalizers. Catalog text is read only after the
candidate and visual scores are frozen; it can raise admission requirements but
cannot regenerate, replace, or edit the frozen value.

\subsection{Counterfactual Visual Candidate Generation}
\label{sec:visual-candidate}

\noindent\textbf{Focus-Zone Distortion.}
Raw question-to-visual attention may cover the complete product rather than the
region that resolves the requested attribute. FZD uses the question-conditioned
attention map $A(I)\in\mathbb R_+^{N_v}$ to produce a localized
counterfactual view $I^{\mathrm{cf}}$, as shown in the FZD block of
Figure~\ref{fig:framework}. It retains the positive attention decrease caused
by weakening that region and normalizes over the visual patches:
\begin{equation}
\begin{aligned}
\Delta_j&=\bigl[A_j(I)-A_j(I^{\mathrm{cf}})\bigr]_+,\\
E_j&=\frac{\Delta_j}{\sum_{r=1}^{N_v}\Delta_r}.
\end{aligned}
\label{eq:fzd}
\end{equation}
The proxy $E$ is attribute-specific because both the probe attention and the
counterfactual intervention are conditioned on the attribute question. It is
computed before any candidate is frozen. A nonpositive normalization term or
inconsistent patch geometry invalidates the route and leads to withholding.

\noindent\textbf{Evidence-Guided Attention Redistribution.}
EGAR strengthens patches supported by $E$ without discarding the model's
original visual distribution. In the EGAR block of Figure~\ref{fig:framework},
let $q_{\mathrm{probe}}$ be the normalized probe map, $q_t$ the visual-key
distribution at a teacher-forced token row, and $z_t$ its attention logits.
On the same patch grid, EGAR computes
\begin{equation}
\begin{aligned}
\gamma&=\gamma_{\max}
\frac{\operatorname{JSD}(q_{\mathrm{probe}},E)}{\log 2},\\
q_{t,j}^{\star}&=(1-\gamma)q_{t,j}+\gamma E_j,\\
z_{t,j}^{\star}&=z_{t,j}+\log\frac{q_{t,j}^{\star}}{q_{t,j}}.
\end{aligned}
\label{eq:egar}
\end{equation}
The Jensen--Shannon divergence makes the correction adaptive while keeping
$\gamma\leq\gamma_{\max}$. The same pair-level $\gamma$ is used for every
teacher-forced row and ontology verbalizer in the fixed attention subset; all
other heads and nonvisual logits remain unchanged. Because $q_t$ and
$q_t^\star$ are normalized, the correction preserves the total exponential
mass over visual keys.

\noindent\textbf{Ontology-Constrained Freezing.}
Each ontology value, the out-of-ontology outcome, and abstention has one frozen
verbalizer. CVE-SAI scores them with length-normalized teacher-forced
log-likelihoods and a temperature-scaled softmax. A fixed ontology order breaks
exact ties, and a canonical value is returned only when its probability reaches
the refusal threshold. The canonical gate then freezes
$(\hat y,p_0,E,\gamma)$, where $p_0$ is the original-image probability of
$\hat y$. Every later view rescores this same value. An out-of-ontology
response, abstention, base fallback, or invalid route is withheld before
auditing.

\subsection{Frozen-Candidate Auditing}
\label{sec:candidate-auditing}

\noindent\textbf{Evidence Necessity and Retention.}
As shown in the auditing block of Figure~\ref{fig:framework}, the evidence mask
$M_E$ is the smallest set of patches whose descending $E$ mass reaches 0.70,
followed by one-patch dilation on the visual grid. The necessity view $I^{-E}$
weakens the masked region, whereas the retention view $I^{+E}$ preserves it and
weakens the complement. Both views reuse the frozen candidate, proxy,
redistribution strength, mask, and scorer. Let $s_J(\hat y)$ be the candidate
probability on view $J$ and $c_J^\dagger$ its top candidate. Writing
$s^-=s_{I^{-E}}(\hat y)$, $s^+=s_{I^{+E}}(\hat y)$, and
$c_+^\dagger=c_{I^{+E}}^\dagger$, the evidence scores are
\begin{equation}
\begin{aligned}
S_{\mathrm{nec}}&=[p_0-s^-]_+,\\
S_{\mathrm{ret}}&=\mathbf{1}[c_+^\dagger=\hat y]
\bigl(1-[p_0-s^+]_+\bigr),\\
S_{\mathrm{evi}}&=\min(S_{\mathrm{nec}},S_{\mathrm{ret}}).
\end{aligned}
\label{eq:evidence-audit}
\end{equation}
Necessity rewards a support decrease after removing the proposed evidence;
retention requires the same candidate to remain top-ranked when that evidence
is preserved.

\noindent\textbf{Nuisance-Transformation Stability.}
Visual support should survive changes unrelated to the target attribute. Let
$I^{(1)},I^{(2)},I^{(3)}$ be the frozen JPEG-compression, linear-RGB
brightness, and outside-mask blur views. With
$s_k=s_{I^{(k)}}(\hat y)$ and top candidate $c_k^\dagger$, the
nuisance-transformation stability score is
\begin{equation}
S_{\mathrm{sta}}=\frac{1}{3}\sum_{k=1}^{3}
\mathbf{1}[c_k^\dagger=\hat y]\bigl(1-|s_k-p_0|\bigr).
\label{eq:stability-audit}
\end{equation}
A top-candidate change contributes zero. The mask and all three transformed
views are mandatory; an invalid element fails the common technical gate.

\noindent\textbf{Candidate-Specific Catalog-Text Conflict.}\par\noindent
After the visual scores are fixed, an ontology-bound parser compares the frozen
candidate with the title and description and returns severity $S_T\in[0,1]$.
Missing text yields $S_T=0$ and is recorded as unobservable. Because the
threshold adjustment below is nonnegative, catalog conflict can only make
admission more demanding; it leaves $\hat y$, $E$, $M_E$, and every visual
score unchanged.

\subsection{Risk-Controlled Index Admission}
\label{sec:risk-admission}

\noindent\textbf{Admission Policy.}
Let $\mathcal B=\{\mathrm{evi},\mathrm{sta}\}$. Each policy
$\theta\in\Theta$ has base thresholds $\tau_b^0$ and nonnegative conflict
penalties $\lambda_b$, and adjusts the two audit thresholds as
\begin{equation}
\tau_b^{\theta}(x)=
\operatorname{clip}\bigl(\tau_b^0+\lambda_bS_T(x),0,1\bigr),
\qquad b\in\mathcal B.
\label{eq:adaptive-threshold}
\end{equation}
A technically valid canonical candidate is admitted only when both
$S_{\mathrm{evi}}$ and $S_{\mathrm{sta}}$ reach their adjusted thresholds;
all other candidates are withheld. Section~\ref{sec:experimental-metrics}
fixes the complete finite grid before risk labels are opened.

\noindent\textbf{Finite-Sample Calibration.}
Let $\mathcal D_R=\mathcal D_{\mathrm{cert}}^{(\mathrm{cal\text{-}risk})}$,
$M=|\Theta|$, and $0<\alpha,\delta<1$. For policy $\theta$, $n_\theta$ is
the number admitted in $\mathcal D_R$ and $k_\theta$ is the number unsafe.
With $\eta=\delta/M$, the Bonferroni-corrected~\cite{Bonferroni1936Teoria}
one-sided Clopper--Pearson bound~\cite{Clopper1934ConfidenceLimits} is
\begin{equation}
U_{\mathrm{CP}}(\theta)=
\begin{cases}
1, & n_\theta=0\ \text{or}\ k_\theta=n_\theta,\\
\mathrm B_{1-\eta}^{-1}(k_\theta+1,n_\theta-k_\theta), & \text{otherwise},
\end{cases}
\label{eq:cp-bound}
\end{equation}
where $\mathrm B_u^{-1}$ is a Beta quantile. The minimum count is fixed as
\begin{equation}
 n_{\min}=\left\lceil\frac{\log(\delta/M)}{\log(1-\alpha)}\right\rceil=172
 \quad(\alpha=\delta=0.05,\ M=324).
 \label{eq:min-admission}
\end{equation}
Thus Figure~\ref{fig:framework}, Eq.~\eqref{eq:risk-constrained-objective}, and
the formal certification rule use the same feasible set:
\begin{equation}
\begin{aligned}
\Theta_{\mathrm{feas}}&=\{\theta\in\Theta:
U_{\mathrm{CP}}(\theta)\leq\alpha,\ n_\theta\geq172\},\\
\theta^\star&=
\operatorname*{arg\,max}^{\mathrm{lex}}_{\theta\in\Theta_{\mathrm{feas}}}
\Pi(\theta).
\end{aligned}
\label{eq:policy-selection}
\end{equation}
The lexicographic tuple $\Pi(\theta)$ orders larger calibration coverage,
smaller upper risk bound, larger minimum admitted score margin, and earlier
preregistered policy ID. This order makes $\theta^\star$ unique. The selected
policy is applied to test without refitting or recertification. If no policy is
feasible, the frozen fallback admits no automatic attributes.

\noindent\textbf{Isolated Index Update.}
Only an admitted canonical value emits a token to the searchable
auto-attribute field. Audit scores and version identifiers remain in a
nonsearchable provenance record. The auto-attribute field is isolated from
merchant-provided fields and can be rolled back without modifying them.

\section{Experimental Setup}
\label{sec:experimental_setup}

\subsection{Datasets and Tasks}
\label{sec:experimental-data}

We construct five visual-attribute tasks from Amazon Berkeley Objects (ABO):
color, pattern, item shape, finish type, and style. Table~\ref{tab:data_statistics}
reports label-blind product-family splits. The test split contains 12,656
eligible product--attribute pairs from 4,000 families. The preselected
certification list fixes one pair per family before annotation or inference;
$\mathcal D_{\mathrm{cert}}$ retains pairs with complete image truth, canonical
or out-of-ontology item truth, and complete evidence annotation. Indeterminate
or incomplete truth and incomplete evidence cause unreplaced attrition, while
visual unanswerability remains eligible. Two annotators label independently and
a third adjudicates disagreements. Localization uses 750 validation pairs and
a disjoint 1,000-pair test set, balanced across the five attributes and with one
pair per family. Controlled retrieval uses 3,712 test certification listings
and 800 query families: 120 attribute queries and 40 controls for validation,
plus 480 attribute queries and 160 controls for test. Predictions, refusals,
and technical failures remain in all applicable denominators.

\begin{table}[t]
  \centering
  \caption{Statistics of the ABO-derived benchmark. Product families define the data splits and resampling units.}
  \label{tab:data_statistics}
  \small
  \setlength{\tabcolsep}{3.5pt}
  \begin{tabular}{lrrrr}
    \toprule
    Split & Families & $\mathcal{D}_{\mathrm{elig}}$
          & $\mathcal{D}_{\mathrm{pre}}$
          & $\mathcal{D}_{\mathrm{cert}}$ \\
    \midrule
    Development & 8,000 & 25,312 & 8,000 & 7,424 \\
    Validation  & 3,000 &  9,492 & 3,000 & 2,784 \\
    Calibration & 5,000 & 15,820 & 5,000 & 4,640 \\
    Test        & 4,000 & 12,656 & 4,000 & 3,712 \\
    \midrule
    Total       & 20,000 & 63,280 & 20,000 & 18,560 \\
    \bottomrule
  \end{tabular}
\end{table}

\subsection{Frozen Models, Parameters, and Metrics}
\label{sec:experimental-metrics}

The primary CVE-SAI backbone is Qwen2.5-VL-3B-Instruct~\cite{Bai2025Qwen25VL};
InternVL3-2B~\cite{Zhu2025InternVL3} is used only in the cross-backbone
analysis. On the primary backbone, $A(I)$ averages question-token attention to
visual tokens over decoder layers 28--35 and all 16 attention heads. FZD
weakens the top 20\% of patches under $A(I)$ with Gaussian blur
$\sigma=12$ pixels, and EGAR uses $\gamma_{\max}=0.35$. The evidence mask uses
the 0.70 cumulative-mass and one-patch-dilation rule in
Section~\ref{sec:candidate-auditing}; necessity and retention use the same
$\sigma=12$ blur. The three nuisance-transformation stability views use JPEG
quality 50, a $1.15\times$ linear-RGB brightness multiplier, and outside-mask
Gaussian blur $\sigma=8$. Candidate probabilities use
$T_{\mathrm{cal}}=0.85$ and refusal threshold $\tau_{\mathrm{ans}}=0.55$.
Catalog fields are Unicode-normalized and matched against frozen ontology
verbalizers before assigning $S_T$. Calibration families are partitioned
30\%/10\%/60\% for probability calibration, policy construction, and
independent risk certification. The policy grid is
$\tau_{\mathrm{evi}}^0\in\{0.05,0.10,0.15,0.20,0.25,0.30\}$,
$\tau_{\mathrm{sta}}^0\in\{0.70,0.75,0.80,0.85,0.90,0.95\}$,
$\lambda_{\mathrm{evi}}\in\{0,0.10,0.20\}$, and
$\lambda_{\mathrm{sta}}\in\{0,0.05,0.10\}$, giving $M=324$ and
$n_{\min}=172$. With $\alpha=\delta=0.05$, calibration selects
$\theta^\star=(0.15,0.85,0.10,0.05)$ in the order
$(\tau_{\mathrm{evi}}^0,\tau_{\mathrm{sta}}^0,
\lambda_{\mathrm{evi}},\lambda_{\mathrm{sta}})$ before test. We report Macro
VAA and Ans.-F1 for attribute inference, Macro Patch AUPRC for evidence
localization, CWC@5\% and UnsafeWWR for admission, and NDCG@10~\cite{Jarvelin2002CumulatedGain}
and UAIE@10 for retrieval.

\subsection{Baselines and Retrieval Protocol}
\label{sec:experimental-baselines}

We compare contrastive encoders CLIP~\cite{Radford2021CLIP},
SigLIP2~\cite{Tschannen2025SigLIP2}, and
FashionCLIP~\cite{Chia2022FashionCLIP}, together with GME~\cite{Zhang2025GME},
MM-Embed~\cite{Lin2025MMEmbed}, InternVL3, Qwen2.5-VL, and MOON. Every method
receives the primary image, attribute question, and frozen ontology, freezes one
candidate, and uses the same family-level risk-selection protocol; external
systems use scalar confidence, while CVE-SAI uses FZD, EGAR, and the four
audits. For localization, we compare FZD with raw question-to-visual attention,
Attention Rollout~\cite{Abnar2020AttentionFlow}, and
Grounding DINO~\cite{Liu2024GroundingDINO}. Retrieval uses Lucene 9.12.1 and BM25~\cite{Robertson2009BM25} with
$k_1=1.2$ and $b=0.75$. Titles and descriptions use the English analyzer;
merchant and auto-attribute fields use keyword tokenization with lowercasing
and accent folding. Query boosts are 2.0 for title, 1.0 for description, and 2.5
for each attribute field. Three-level qrels are fixed before automatic
attributes are generated: grade 2 requires product-type relevance and an exact
trusted target-attribute match, grade 1 denotes product-type relevance without
a verified exact match, and grade 0 denotes nonrelevance. Each automatic run is
paired with the same system's No-Auto run, which differs only by leaving the
auto-attribute field empty. UAIE@10 counts top-10 positions newly occupied by
grade-0 products through unsafe automatic values. The 160 test controls contain
no target-attribute cue, so the query builder omits both merchant-attribute and
auto-attribute clauses. Ties use ascending listing ID.

\section{Results and Analysis}
\label{sec:results}

We first examine attribute inference and evidence localization, then test whether these gains yield broader certified admission and stronger retrieval. Further analyses cover individual components, attributes, backbones, risk budgets, and withholding outcomes.

\subsection{Attribute Inference and Evidence Localization}
\label{sec:results-upstream}

Table~\ref{tab:front_end_results}(a) compares image-only attribute inference under the shared image-only input protocol. On 12,656 eligible test pairs, CVE-SAI obtains 72.36\% Macro VAA and 84.47\% Ans.-F1, outperforming the strongest baseline, MOON, by 2.95 and 2.75 percentage points. These differences correspond to relative improvements of 4.25\% and 3.37\%, respectively. The simultaneous gains indicate that the front end improves exact ontology-value prediction without trading away recognition of image-unanswerable cases. Because every eligible pair remains in the denominator, fallback and failed-inference outcomes directly lower both scores rather than being removed from evaluation.

\begin{table}[t]
  \centering
  \caption{Results on the ABO-derived test sets: (a) attribute inference and (b) evidence localization. Values are percentages; best and second-best results are shown in bold and underlined.}
  \label{tab:front_end_results}
  \small
  \setlength{\tabcolsep}{4.2pt}
  \renewcommand{\arraystretch}{1.02}
  \begin{tabular}{@{}lcc@{}}
    \toprule
    \multicolumn{3}{c}{\textbf{(a) Attribute inference}} \\
    \midrule
    Method & Macro VAA$\uparrow$ & Ans.-F1$\uparrow$ \\
    \midrule
    SigLIP2 & 57.26 & 72.44 \\
    InternVL3 & 65.42 & 78.88 \\
    Qwen2.5-VL & 67.84 & 80.31 \\
    MOON & \underline{69.41} & \underline{81.72} \\
    CVE-SAI Front End & \textbf{72.36} & \textbf{84.47} \\
    \bottomrule
  \end{tabular}
  \par\vspace{5pt}
  \setlength{\tabcolsep}{5pt}
  \begin{tabular}{@{}lc@{}}
    \toprule
    \multicolumn{2}{c}{\textbf{(b) Evidence localization}} \\
    \midrule
    Method & Patch AUPRC$\uparrow$ \\
    \midrule
    Grounding DINO & 30.84 \\
    Attention Rollout & \underline{38.27} \\
    FZD & \textbf{49.16} \\
    \bottomrule
  \end{tabular}
\end{table}

Table~\ref{tab:front_end_results}(b) evaluates evidence localization with Macro Patch AUPRC. FZD reaches 49.16\%, improving over Attention Rollout by 10.89 points and Grounding DINO by 18.32 points, or 28.46\% and 59.40\% in relative terms. Generic object localization tends to cover the complete product, although a queried attribute may depend on a small pattern, material, or surface region. FZD instead assigns patch importance according to the positive decrease in question-conditioned visual attention caused by the focus-zone intervention. As shown in the front-end path of Figure 1, the resulting attribute-specific visual-dependence proxy is computed before candidate freezing and guides EGAR in refining ontology-constrained scores. The gain in Table~\ref{tab:front_end_results}(a) is therefore accompanied by a substantially more selective estimate of where the supporting visual information lies.

Answerability and localization capture different failure modes. Ans.-F1 penalizes a system that answers when the primary image is insufficient or refuses when the attribute is visible, whereas Patch AUPRC tests whether the estimated support coincides with the annotated attribute region. CVE-SAI improves both quantities, so its higher answer accuracy is not produced by answering more pairs without regard to visual support. This distinction is important for indexing: a correct-looking value still requires localized evidence before it can become a persistent retrieval signal.

\subsection{Risk-Controlled Index Admission}
\label{sec:results-admission}

\begin{figure}[t]
  \centering
  \includegraphics[width=\columnwidth]{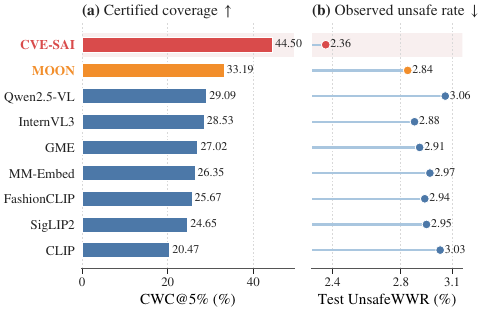}
  \caption{Risk-controlled index admission under the shared 5\% certification protocol: (a) CWC@5\% and (b) observed test unsafe-admission rate. Higher CWC@5\% and lower UnsafeWWR are better.}
  \label{fig:certified-admission}
\end{figure}

Table~\ref{tab:certified_admission} and Figure~\ref{fig:certified-admission} report risk-controlled index admission under the shared 5\% unsafe-admission budget. The comparison spans contrastive encoders, product-specific representations, universal multimodal embedders, and generative multimodal models under the common protocol described in Section~\ref{sec:experimental-baselines}.

\begin{table}[t]
  \centering
  \caption{Risk-controlled index admission on 3,712 test certification pairs. CWC@5\% uses policies selected under the shared 5\% risk protocol; UnsafeWWR is observed on test. Values are percentages; best and second-best results are shown in bold and underlined.}
  \label{tab:certified_admission}
  \small
  \setlength{\tabcolsep}{3.2pt}
  \renewcommand{\arraystretch}{1.01}
  \begin{tabular}{@{}lcc@{}}
    \toprule
    Method & CWC@5\%$\uparrow$ & UnsafeWWR$\downarrow$ \\
    \midrule
    CLIP & 20.47 & 3.03 \\
    SigLIP2 & 24.65 & 2.95 \\
    FashionCLIP & 25.67 & 2.94 \\
    GME & 27.02 & 2.91 \\
    MM-Embed & 26.35 & 2.97 \\
    InternVL3 & 28.53 & 2.88 \\
    Qwen2.5-VL & 29.09 & 3.06 \\
    MOON & \underline{33.19} & \underline{2.84} \\
    \midrule
    CVE-SAI & \textbf{44.50} & \textbf{2.36} \\
    \bottomrule
  \end{tabular}
\end{table}

For CVE-SAI, the frozen $\theta^\star=(0.15,0.85,0.10,0.05)$ admits 1,184 risk-certification candidates, of which 32 are unsafe. Its empirical UnsafeWWR is 2.70\%, while the selector-corrected one-sided bound is 4.830\%. The 2.13-point gap reflects finite-sample uncertainty and simultaneous search over the prespecified $M=324$ policies. The bound remains 0.170 points below the 5\% limit; this same policy is then applied to test without refitting or recertification.

After transfer to the test population, the observed 2.36\% unsafe rate is 0.34 points below the calibration empirical rate and 2.47 points below its corrected upper bound. The same frozen thresholds produce these test outcomes; no test-specific operating point is selected. The calibration result and the subsequent test count therefore describe one policy from selection through test evaluation.

CVE-SAI attains 44.50\% CWC@5\%, compared with 33.19\% for MOON. This is an 11.31-point absolute gain and a 34.08\% relative increase over the strongest external system. Its observed UnsafeWWR simultaneously decreases from 2.84\% to 2.36\%, a 0.48-point absolute and 16.90\% relative reduction. Thus, the additional coverage is not obtained by admitting a larger unsafe fraction.

The underlying counts make this difference concrete. On 3,712 test certification pairs, CVE-SAI admits 1,652 candidates, including 1,613 that are both factually correct and visually supported and 39 that are unsafe. MOON admits 1,232 candidates, with 1,197 safe and 35 unsafe admissions. CVE-SAI therefore adds 420 admitted attributes: 416 safe and four unsafe. Safe values account for 99.05\% of this incremental set, and the total number of safe admitted attributes increases by 34.75\%. These gains directly enlarge the searchable attribute inventory while preserving a lower unsafe proportion.

When normalized by the complete test certification population, CVE-SAI supplies 43.45 safe admitted values per 100 eligible certification pairs, compared with 32.25 for MOON. The resulting 11.21-point gain closely tracks the 11.31-point CWC improvement because nearly all additional admissions are safe. At the same time, the unsafe count rises by only four while the admitted set grows by 420. The selector therefore uses the available risk budget primarily to recover useful attributes rather than to relax support requirements uniformly.

Scalar confidence alone cannot distinguish a plausible canonical value from one that lacks attribute-relevant visual support. CVE-SAI evaluates evidence necessity, evidence retention, nuisance-transformation stability, and candidate-specific catalog-text conflict after candidate freezing. The first two audits test whether removing or retaining the localized region changes support in the expected direction; the nuisance-transformation stability audit rejects values that react excessively to nuisance variation; and the conflict audit raises the admission requirement when catalog text disagrees with the frozen value. Their combination separates visual support from answer confidence and explains why coverage can rise while observed unsafe admission falls.

The external front ends all use the same risk-selection procedure, so calibration alone cannot account for the gap in Table~\ref{tab:certified_admission}. Their CWC@5\% values range from 20.47\% to 33.19\%, even though their observed UnsafeWWR values remain within a narrow 2.84\%--3.06\% interval. CVE-SAI moves beyond this cluster in both directions, reaching 44.50\% coverage at 2.36\% observed unsafe admission. The evidence audits provide additional ordering information among candidates whose scalar scores alone offer similar risk--coverage choices.

The minimum-admission condition prevents a policy from appearing certifiable by writing only a handful of easy cases. The selected policy admits 1,184 risk-certification candidates, well above the required 172, so its operating point is not determined by the count floor. Its 4.830\% upper bound instead reflects the 32 observed unsafe admissions, the admitted sample size, and the Bonferroni correction over the frozen policy family. CWC@5\% therefore evaluates the coverage of one statistically eligible policy rather than a threshold chosen retrospectively from test outcomes.

\subsection{Retrieval Effectiveness}
\label{sec:results-retrieval}

\begin{figure}[t]
  \centering
  \includegraphics[width=\columnwidth]{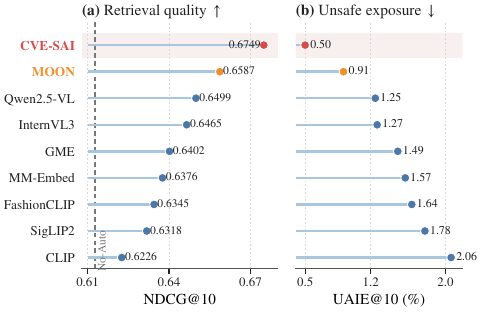}
  \caption{Controlled retrieval comparison: (a) NDCG@10 with the No-Auto reference and (b) unsafe auto-induced exposure among automatic-admission systems. Higher NDCG@10 and lower UAIE@10 are better.}
  \label{fig:retrieval-effectiveness}
\end{figure}

Table~\ref{tab:retrieval_results} and Figure~\ref{fig:retrieval-effectiveness} report the retrieval effect of the attributes admitted in Table~\ref{tab:certified_admission}. Each policy supplies canonical values to the same isolated auto-attribute field under the fixed Lucene configuration.

\begin{table}[t]
  \centering
  \caption{Controlled Lucene retrieval on 480 attribute-bearing test queries using the admitted values from Table~\ref{tab:certified_admission}. A dash denotes no automatic-attribute exposure; best and second-best results are shown in bold and underlined.}
  \label{tab:retrieval_results}
  \small
  \setlength{\tabcolsep}{3.2pt}
  \renewcommand{\arraystretch}{1.01}
  \begin{tabular}{@{}lcc@{}}
    \toprule
    Method & NDCG@10$\uparrow$ & UAIE@10 (\%)$\downarrow$ \\
    \midrule
    No-Auto & 0.6128 & -- \\
    CLIP & 0.6226 & 2.06 \\
    SigLIP2 & 0.6318 & 1.78 \\
    FashionCLIP & 0.6345 & 1.64 \\
    GME & 0.6402 & 1.49 \\
    MM-Embed & 0.6376 & 1.57 \\
    InternVL3 & 0.6465 & 1.27 \\
    Qwen2.5-VL & 0.6499 & 1.25 \\
    MOON & \underline{0.6587} & \underline{0.91} \\
    \midrule
    CVE-SAI & \textbf{0.6749} & \textbf{0.50} \\
    \bottomrule
  \end{tabular}
\end{table}

CVE-SAI reaches 0.6749 NDCG@10. This exceeds MOON by 0.0162, or 2.46\%, and the paired No-Auto reference by 0.0621, or 10.13\%. Within each paired run, the corpus, analyzers, BM25 parameters, query, qrels, boosts, and tie breaking are fixed; the only changed input is that system's admitted auto-attribute tokens. Under this controlled protocol, the within-system ranking difference is therefore attributable to those tokens rather than to a changed retrieval pipeline.

The progression among the strongest systems further connects admission quality to retrieval utility. Qwen2.5-VL reaches 29.09\% CWC@5\% and 0.6499 NDCG@10; MOON increases these values to 33.19\% and 0.6587; CVE-SAI reaches 44.50\% and 0.6749. Unsafe exposure decreases over the same sequence from 1.25\% to 0.91\% and then 0.50\%. Thus, the ranking gain is associated with a larger pool of admitted attributes whose unsafe share is lower, rather than with indiscriminate expansion of the auto-attribute field.

CVE-SAI also obtains the lowest UAIE@10 among automatic-admission systems. Its value of 0.50\% is 0.41 percentage points below MOON, a 45.05\% relative reduction. This result complements NDCG@10: the system retrieves more relevant products while exposing fewer nonrelevant products through unsafe automatic attributes. The admission gain in Table~\ref{tab:certified_admission} therefore survives downstream use, improving ranking quality instead of merely increasing the number of indexed values. The 160 control queries activate neither attribute field; every paired ranking is unchanged, with $\Delta$NDCG@10$=0.0000$ and UAIE@10$=0$. The retrieval effect is therefore confined to queries that contain the evaluated attribute cue. The attribute and control queries serve complementary roles: attribute queries activate the exact-match clauses and test the intended retrieval route, whereas controls retain the same corpus and lexical pipeline without an attribute cue. Their zero paired differences rule out score changes caused by analyzer drift, index construction, or listing-ID tie handling. The NDCG@10 and UAIE@10 changes are observed only when the auto-attribute field is eligible to contribute.

\subsection{Component Analysis}
\label{sec:results-components}

The variants in Table~\ref{tab:component_ablation} correspond to the visual-inference and auditing components shown in Figure~\ref{fig:framework}. Replacing FZD with raw attention reduces CWC@5\% from 44.50\% to 35.64\% and NDCG@10 from 0.6749 to 0.6578, absolute drops of 8.86 points and 0.0171. Removing EGAR lowers the same metrics to 38.36\% and 0.6632, drops of 6.14 points and 0.0117. FZD thus supplies a more useful attribute-specific visual-dependence proxy than raw attention, while EGAR converts that proxy into better ontology-constrained scores before freezing the candidate.

\begin{table}[t]
  \centering
  \caption{Component analysis under the shared 5\% risk protocol. Each variant is recalibrated independently; best and second-best results are shown in bold and underlined.}
  \label{tab:component_ablation}
  \footnotesize
  \setlength{\tabcolsep}{3.0pt}
  \renewcommand{\arraystretch}{1.01}
  \begin{tabular}{@{}lcc@{}}
    \toprule
    Variant & CWC@5\%$\uparrow$ & NDCG@10$\uparrow$ \\
    \midrule
    Raw attention for FZD & 35.64 & 0.6578 \\
    w/o attention redistribution & 38.36 & 0.6632 \\
    Necessity only & 31.84 & 0.6582 \\
    Retention only & 33.62 & 0.6604 \\
    w/o nuisance-transformation stability & 38.79 & 0.6658 \\
    w/o conflict audit & \underline{41.11} & \underline{0.6687} \\
    Full CVE-SAI & \textbf{44.50} & \textbf{0.6749} \\
    \bottomrule
  \end{tabular}
\end{table}

Necessity alone trails the full system by 12.66 coverage points and 0.0167 NDCG@10, while retention alone trails it by 10.88 points and 0.0145. Each audit observes only one side of the intervention: necessity measures the effect of removing evidence, whereas retention checks whether the retained region preserves the candidate. Their joint use provides a stronger admission signal than either direction separately.

Removing nuisance-transformation stability decreases CWC@5\% by 5.71 points and NDCG@10 by 0.0091. Removing candidate-specific catalog-text conflict produces smaller but consistent drops of 3.39 points and 0.0062. The latter audit acts only on admission, leaving the frozen visual value unchanged. Finally, an empirical threshold without finite-sample certification reaches 50.03\% coverage but also 5.82\% UnsafeWWR. The 5.53-point coverage increase over full CVE-SAI comes with an unsafe rate 0.82 points above the budget. Finite-sample calibration is therefore essential to the observed balance between admitted coverage and unsafe admission.

The ordering of the ablations separates upstream evidence quality from downstream admission checks. Raw attention causes the largest drop among the two front-end variants, followed by removing EGAR, which agrees with the localization advantage in Table~\ref{tab:front_end_results}(b). Among the audits, using only one counterfactual direction loses more coverage than removing either nuisance-transformation stability or candidate-specific catalog-text conflict from the complete selector. Necessity and retention therefore form the core visual-support test, while these two audits recover complementary failure modes that remain after localized evidence has been assessed.

Every structural variant is recalibrated independently on the same family-level splits, rather than inheriting thresholds optimized for the full system. The comparison therefore asks whether each altered score set can support a high-coverage policy under the same 5\% risk protocol. The aligned reductions in CWC@5\% and NDCG@10 show that upstream evidence quality and downstream audits affect the usefulness of the resulting ordering even after each variant receives its own best certifiable operating point.

\subsection{Cross-Attribute and Backbone Robustness}
\label{sec:results-robustness}

Across the five predefined attributes, observed coverage ranges from 36.30\% for style to 50.26\% for color, a span of 13.96 points. Observed UnsafeWWR ranges from 1.28\% to 3.89\%, and even the largest slice-level value remains 1.11 points below 5\%. All slices inherit the same globally selected policy, with no attribute-specific threshold. Color therefore offers more admissible visual evidence than style under the shared rule, while the unsafe proportions remain consistently low across attributes.

The 2.61-point spread in observed UnsafeWWR is substantially smaller than the 13.96-point coverage spread. The shared policy consequently adapts mainly through the number of candidates satisfying its evidence conditions, not through large attribute-wise changes in the unsafe fraction. For a catalog containing heterogeneous attributes, slices that satisfy the common evidence criteria more often contribute more indexable values, while the same rule withholds a larger share of the remaining slices.

Qwen2.5-VL-3B-Instruct is the primary backbone. On the secondary InternVL3-2B backbone, applying the CVE-SAI front end raises Macro VAA from 65.42\% to 69.81\%, an absolute gain of 4.39 points and a relative gain of 6.71\%. The corresponding gain with Qwen2.5-VL is 4.52 points, differing by only 0.13 points. InternVL3 with CVE-SAI reaches 0.6691 NDCG@10, 0.0226 above the unmodified InternVL3 retrieval result and only 0.0058 below the primary CVE-SAI system. The candidate-generation and admission design therefore transfers across the two evaluated multimodal backbones.

\subsection{Risk-Budget Sensitivity}
\label{sec:results-risk-sensitivity}

The risk budget controls how much certified coverage the selector can retain. Tightening the budget to 1\% yields 31.25\% test coverage, the primary 5\% budget yields 44.50\%, and relaxing it to 10\% yields 64.01\%. Coverage increases by 13.25 points from 1\% to 5\% and by a further 19.51 points from 5\% to 10\%, for a total span of 32.76 points. The 5\% operating point retains substantially more coverage than the 1\% setting while its selected policy has a 4.830\% calibration bound. This operating point is used for every main comparison and all downstream retrieval runs.

Measured per additional percentage point of risk budget, the first interval recovers 3.31 coverage points and the second recovers 3.90. The selector therefore has a broad set of candidates near the stricter evidence thresholds, but admitting them requires a correspondingly looser certified risk level. Fixing the operating point before the test evaluation prevents this trade-off from being chosen retrospectively from test coverage.

The sensitivity curve is a genuine risk--coverage trade-off rather than a cost-free gain. The 10\% setting admits many candidates that fail the stricter evidence requirements of the primary policy, but it also permits a larger certified unsafe-admission budget. Conversely, the 1\% setting withholds additional visually plausible values to obtain a tighter guarantee. The 5\% operating point is therefore interpreted only as the prespecified deployment budget used for the main study, not as evidence that the same threshold is optimal for every catalog or application.

\subsection{Withholding Analysis}
\label{sec:results-withholding}

The frozen policy withholds 2,060 of the 3,712 test candidates, or 55.50\%. Insufficient evidence or nuisance-transformation stability accounts for 1,301 cases (63.16\%), making it the dominant reason. Another 376 candidates (18.25\%) fall outside the frozen ontology, and 180 (8.74\%) are withheld after candidate-specific catalog-text conflict raises the admission requirement. These three substantive outcomes comprise 90.15\% of all withheld cases and prevent unsupported, noncanonical, or conflict-sensitive values from becoming searchable attributes.

The remaining 167 cases (8.11\%) have unusable FZD or mask outputs, and 36 (1.75\%) encounter another technical failure. Together they constitute 9.85\% of withheld candidates. Catalog text remains asymmetric in this analysis: it can make admission more demanding but cannot replace the frozen visual value. The distribution of outcomes shows that withholding is driven mainly by evidence quality and ontology fit rather than by technical attrition.

Evidence or nuisance-transformation stability insufficiency alone is 6.41 times as frequent as the two technical categories combined. Out-of-ontology values and catalog conflicts account for another 556 cases, or 26.99\% of all withholding. These counts locate the principal constraint on coverage in ambiguous visual support and catalog compatibility, while implementation failures form a comparatively small tail. The withheld set therefore reflects the intended selectivity of the admission stage rather than a large loss of otherwise usable candidates to system errors.

The three final dispositions exhaust the 3,712-pair certification population: 1,613 safe admissions, 39 unsafe admissions, and 2,060 withheld candidates. Per 100 certification pairs, the policy contributes 43.45 safe indexed values, admits 1.05 unsafe values, and withholds 55.50 candidates. This partition exposes the operational trade-off directly: coverage is limited chiefly by withholding, while unsafe additions remain a small part of the resulting auto-attribute field.

\section{Conclusions and Future Work}
\label{sec:conclusion}

We formulate risk-controlled selective product attribute indexing and propose
CVE-SAI, which separates attribute inference from persistent index
admission. FZD constructs an attribute-specific visual-dependence proxy, EGAR
uses it to refine ontology-constrained scoring, and the canonical candidate is
frozen before evidence necessity, evidence retention, nuisance-transformation
stability, and candidate-specific catalog-text conflict audits. A finite policy
family and independent one-sided calibration select a unique admission rule
under a 5\% unsafe-admission budget, after which admitted values enter only a
reversible auto-attribute field. Across five ABO attributes, CVE-SAI improves attribute inference and evidence localization, achieves the highest CWC@5\% with
the lowest observed UnsafeWWR, and provides the highest NDCG@10 with the lowest
UAIE@10 among automatic-admission systems. Future work will extend the framework
to multi-image products, evolving ontologies, and cross-catalog calibration
while preserving the separation among inference, auditing, and admission.

\bibliographystyle{unsrtnat}
\bibliography{references}

\end{document}